\documentclass[10pt,twocolumn,letterpaper]{article}

\usepackage[pagenumbers]{cvpr} 

\usepackage{footnote}
\usepackage{threeparttable}

\definecolor{cvprblue}{rgb}{0.21,0.49,0.74}
\usepackage[pagebackref,breaklinks,colorlinks,allcolors=cvprblue]{hyperref}

\def\paperID{3} 
\def\confName{CVPR}
\def\confYear{2026}

\title{4D Radar Meets LiDAR and Camera: Cooperative Perception under Adverse Weather}

\author{%
  Melih Yazgan\textsuperscript{1,2}\textsuperscript{†}\quad
  Iramm Hamdard\textsuperscript{1,2}\textsuperscript{†}\quad
  Qiyuan Wu\textsuperscript{2}\quad
  J.~Marius Zoellner\textsuperscript{1,2}\\
  \textsuperscript{1}FZI Research Center for Information Technology, 
  \textsuperscript{2}Karlsruhe Institute of Technology\\
  \footnotesize\texttt{last.name@fzi.de}\\\footnotesize\textsuperscript{†}These authors contributed equally.
}

\begin{document}
\maketitle
\begin{abstract}
Cooperative perception is important for autonomous driving but remains fragile when cameras and LiDAR degrade in adverse weather. We address this challenge by integrating 4D imaging radar as a weather-robust modality into collaborative perception and introducing a Doppler-guided spatial attention mechanism for multi-agent fusion. Our approach extends two representative backbones: a radar-camera pipeline where radar substitutes LiDAR, and a LiDAR-radar pipeline where radar complements LiDAR. To support evaluation, we release radar-augmented benchmarks, OPV2V-R and Adver-City-R, with physics-based LiDAR degradation. Experiments show strong robustness gains in fog and rain, including substantial improvements when radar replaces degraded LiDAR. Additional validation on MAN TruckScenes demonstrates transfer beyond simulation. Overall, our results highlight 4D imaging radar as a robust modality for all-weather collaborative perception. Dataset and code are available at: \url{https://url.fzi.de/SlimComm}.
\end{abstract}    
\section{Introduction}
The safe deployment of autonomous vehicles hinges on reliable perception. While autonomous driving has made remarkable progress in recent years, particularly in environmental perception \cite{Liu_2025,wu2025foundationmodelsautonomousdriving,liu2024surveyautonomousdrivingdatasets}, failures in perception directly translate into safety risks. Robust perception remains a prerequisite for large-scale deployment \cite{sun2020scalabilityperceptionautonomousdriving}, yet adverse weather such as rain and fog can severely degrade camera- and LiDAR-based systems \cite{wang2020fusion}. Cameras suffer from reduced visibility, while LiDAR suffers from scattering and attenuation, leading to degraded detection and tracking performance.  

To mitigate these vulnerabilities, modern autonomous platforms adopt multi-sensor fusion strategies that exploit complementary strengths across modalities \cite{wang2020fusion,wevj2025,harley2022simple}. Among them, four-dimensional (4D) imaging radar has attracted growing interest because it remains robust in poor visibility and directly provides Doppler velocity measurements \cite{peng20254dmmwaveradarsensing, zhang2023perception}. Its resilience makes 4D radar a promising complementary modality for all-weather perception, yet it remains underexplored in collaborative frameworks.  
\begin{figure}[t!]
    \centering
    \includegraphics[width=0.85\linewidth]{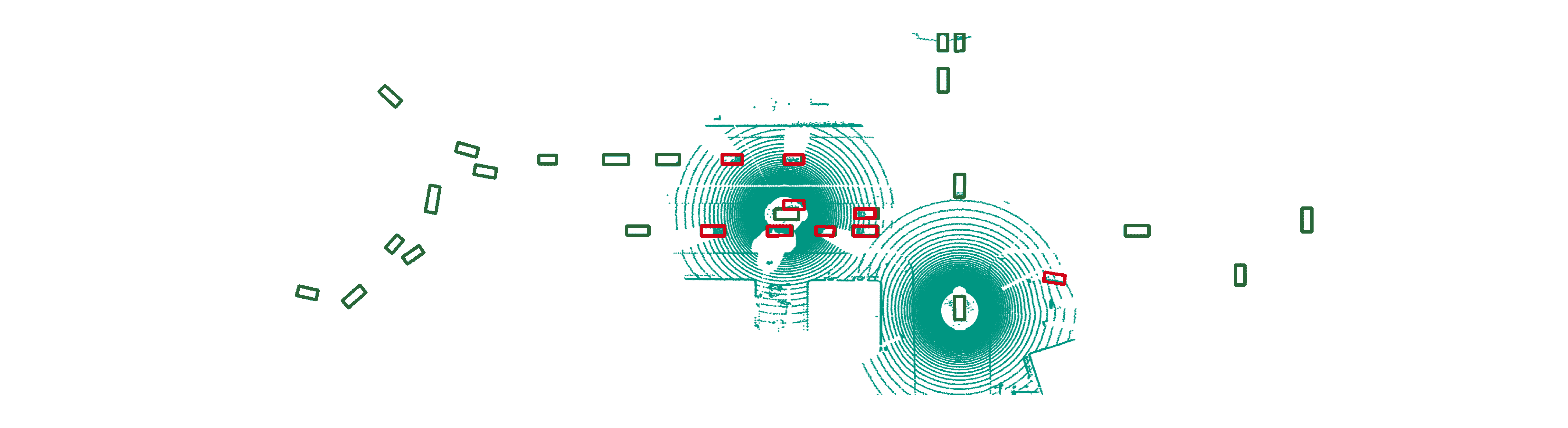}
    \includegraphics[width=0.85\linewidth]{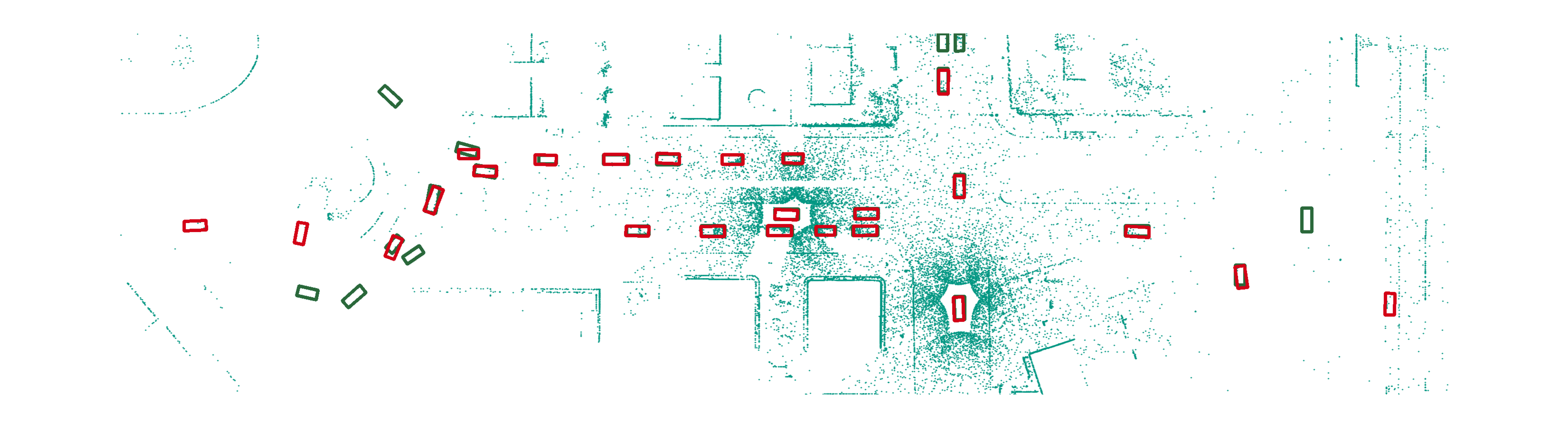}
    \caption{Impact of adverse weather on LiDAR and radar perception. 
    Top: LiDAR point clouds degrade heavily under fog and rain, losing density and range. 
    Bottom: Radar remains stable, with Doppler cues highlighting movers. 
    This complementary robustness motivates our radar-augmented collaborative fusion. (Simulated via CARLA~\cite{carla} and the LISA~\cite{kilic2021lidar} framework).}
    \label{fig:degredation}
\end{figure}
A parallel trend is collaborative perception, where connected vehicles and infrastructure share sensor information to overcome occlusion and extend perceptual range. Recent surveys and benchmarks highlight its potential and open challenges \cite{huang2024v2xcooperativeperceptionautonomous,huang2024v2x,Yao_2024}. However, the majority of collaborative perception methods have been evaluated predominantly under clear-weather conditions, with limited systematic treatment of multi-modal sensor degradation under adverse weather~\cite{yazgan2024real}. LiDAR, the dominant modality in collaborative frameworks, is also fragile in adverse weather, as illustrated in Fig.~\ref{fig:degredation}.
Progress is further bottlenecked by the absence of suitable datasets. Widely used resources such as OPV2V \cite{xu2022opencood}, Adver-City \cite{karvat2025advercityopensourcemultimodaldataset}, and DAIR-V2X \cite{dair-v2x} each miss essential requirements such as synchronized 4D radar, realistic weather degradation, or full 360° coverage. Reviews of collaborative-perception datasets confirm that no public benchmark currently satisfies the combined need for 4D radar, 360° layouts, and physics-based weather simulation \cite{YazganCoPeData,wang2025collaborativeperceptiondatasetsautonomous}. SlimComm~\cite{yazgan2025slimcomm} previously integrated radar into OPV2V and Adver-City and benchmarked LiDAR-radar collaborative perception, but left weather-induced sensor degradation unaddressed.
\noindent Our contributions are threefold:
\begin{enumerate}
    \item \textbf{Radar-augmented collaborative fusion:} We integrate 4D radar into two representative collaborative baselines: substituting LiDAR in BM2CP to form a radar-camera pipeline, and complementing LiDAR in Where2comm to form a LiDAR-radar pipeline.
    \item \textbf{Doppler-guided spatial attention:} We introduce an attention mechanism that leverages radar Doppler velocity to create a dynamic Bird's-Eye-View (BEV) mask, effectively emphasizing moving objects across multi-agent features.
    \item \textbf{Adverse-weather benchmarks and evaluation:} We release OPV2V-R and Adver-City-R with physics-based weather degradation. We demonstrate strong robustness against severe weather, communication latency, and spatial misalignment, and validate our modules on the real-world MAN TruckScenes~\cite{fent2024mantruckscenesmultimodaldataset} single-vehicle dataset to confirm transferability beyond simulation.
\end{enumerate}
\section{Related Work}

\subsection{Radar Fusion in Single-Vehicle Perception}
Autonomous vehicles typically rely on cameras, LiDAR, and radar as core exteroceptive sensors. Cameras provide rich semantics but degrade under poor visibility, while LiDAR delivers precise geometry but suffers from scattering and attenuation in rain and fog~\cite{zhang2023perception,teufel2023weather}. Radar, by contrast, is inherently resilient to visibility degradation and directly provides Doppler velocity cues. The emergence of 4D radar has therefore motivated a range of fusion methods that exploit radar either in combination with cameras, LiDAR, or as a standalone modality.

Early approaches such as CRFNet~\cite{nobis2019crfnet} and RAMP-CNN~\cite{ramp_cnn} integrated radar detections with images, while CenterFusion~\cite{nabati2021centerfusion}, RCFusion~\cite{xu2022rcfusion}, and CRAFT~\cite{Kim2022CRAFT} demonstrated improved detection under challenging conditions. More recent works leverage 4D radar directly: InterFusion~\cite{wang2022interfusion} proposed interaction-based LiDAR-radar fusion, LXL~\cite{xiong2024lxl} excluded LiDAR entirely with a radar occupancy-assisted depth strategy, and MSSF~\cite{mssf2024} introduced a multi-stage radar-camera sampling framework. RadarPillars~\cite{radar_pillar} further advanced radar-only detection by enriching velocity features with intra-pillar offsets and applying radar-tailored attention.

Taken together, these approaches confirm 4D radar as a weather-resilient and cost-effective modality for perception. However, they remain confined to the \emph{single-vehicle} setting and do not address multi-agent collaboration.

\subsection{Collaborative Perception under Adverse Weather}
Collaborative perception extends perceptual range and mitigates occlusions by allowing connected vehicles to exchange features. Most prior works have focused on LiDAR-camera pipelines~\cite{tsakmakopoulou2023v2v,jiang2023weather,li2025dg}, showing that collaboration improves robustness but still inherits LiDAR’s fragility in adverse weather. Jiang et al.~\cite{jiang2023weather} addressed uncertainty in rain via probabilistic feature aggregation, while Li et al.~\cite{li2025dg} proposed domain generalization to unseen weather. Tsakmakopoulou and Moustakas~\cite{tsakmakopoulou2023v2v} showed that V2V communication helps mitigate fog degradation in cooperative pipelines by integrating their proposed method on top of S-AdaFusion~\cite{qiao2023adaptive}. Complementing these, Wang et al.~\cite{wang2024sensors} surveyed adverse-weather cooperative strategies, emphasizing the need for efficiency under communication constraints. Taking a different approach, Huang et al.~\cite{huang2024v2x} introduced AttFuse w/MDD, a denoising diffusion model that uses weather-robust 4D radar features to guide the reconstruction of noisy LiDAR features. While advanced techniques like denoising diffusion show promise for improving accuracy, they can also introduce significant computational overhead. In contrast, our work prioritizes an efficient solution, and no prior collaborative perception method has explicitly leveraged radar Doppler cues to guide attention across agents in a lightweight manner.

\subsection{Datasets for Collaborative Radar Perception}
Progress in collaborative perception relies heavily on dedicated datasets. Large-scale single-vehicle datasets such as nuScenes~\cite{caesar2020nuscenes}, K-Radar~\cite{kim2022kradar}, and MAN TruckScenes~\cite{fent2024mantruckscenesmultimodaldataset} include radar and adverse-weather conditions but lack cooperative settings. Collaborative datasets such as OPV2V~\cite{xu2022opencood}, DAIR-V2X~\cite{dair-v2x}, and Adver-City~\cite{karvat2025advercityopensourcemultimodaldataset} support multi-agent fusion, but OPV2V lacks realistic weather, DAIR-V2X has restricted modality coverage, and Adver-City does not model LiDAR degradation. TUMTraf-V2X~\cite{Zimmer_2024_CVPR} provides real-world V2X cooperative perception with camera and LiDAR, but does not include radar as a cooperative modality.

Radar-focused cooperative datasets remain limited. V2X-R~\cite{huang2024v2x} provides simulated radar perception but without full 360° coverage and with Doppler encoded only as a boolean flag. V2X-Radar~\cite{yang2025v2xradarmultimodaldataset4d} introduces real-world radar data, but its public release omits cooperative elements and suffers from fidelity issues in ego-velocity and Doppler. SlimComm~\cite{yazgan2025slimcomm} contributed radar-augmented versions of OPV2V and Adver-City, enabling early LiDAR-radar cooperative benchmarks, but without modeling adverse weather.  

\textbf{Summary.} Existing works confirm the promise of radar-camera fusion for single-vehicle perception, with specialized designs such as RadarPillars introducing velocity offsets, pillar-level attention, and uniform backbone scaling. However, these approaches are tailored to radar-only pipelines and do not address collaboration. Collaborative methods, by contrast, still rely primarily on LiDAR and remain fragile in adverse weather. LiDAR-radar cooperative fusion has been attempted in SlimComm~\cite{yazgan2025slimcomm}, yet without Doppler-guided attention and without systematic weather modeling. Our work addresses these gaps by (i) proposing the Doppler-guided attention for collaborative radar fusion, applicable to both radar-camera and LiDAR-radar backbones, and (ii) extending the SlimComm datasets with physics-based weather degradation, resulting in OPV2V-R and Adver-City-R benchmarks for all-weather collaborative perception.

\section{Methodology}
\begin{figure*}[ht!]
      \centering
      \includegraphics[width=0.90\linewidth]{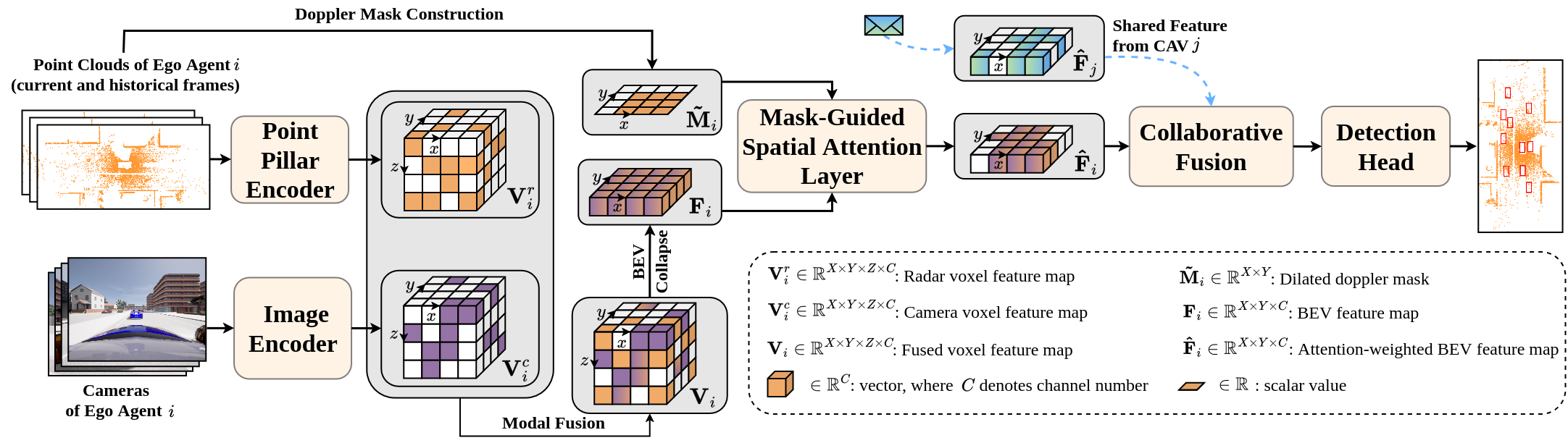}
      \caption{Overview of the proposed architecture, illustrated on BM2CP where radar replaces LiDAR. Radar features are encoded via a velocity-conditioned PillarVFE, and a Doppler-derived mask guides residual spatial attention in BEV space. In Where2comm, the same radar modules are used, but radar complements LiDAR features rather than substituting them.}
      \label{fig:architecture}
\end{figure*}

Our approach extends two representative collaborative perception backbones: 
BM2CP~\cite{zhao2023bm2cp}, which fuses LiDAR and camera, and 
Where2comm~\cite{hu2022where2comm}, which is LiDAR-only and communication-efficient. 
We integrate 4D radar into both: 
(i) in BM2CP, radar substitutes LiDAR, forming a radar-camera pipeline, and 
(ii) in Where2comm, radar complements LiDAR, forming a LiDAR-radar pipeline. 
In both cases, we further introduce a Doppler-guided spatial attention mechanism that 
emphasizes moving objects while preserving static context.~\cref{fig:architecture} illustrates our design on BM2CP; integration into Where2comm follows the same principles but with radar as an additional modality.

\begin{figure}[b!]
    \centering
    \includegraphics[width=0.9\linewidth]{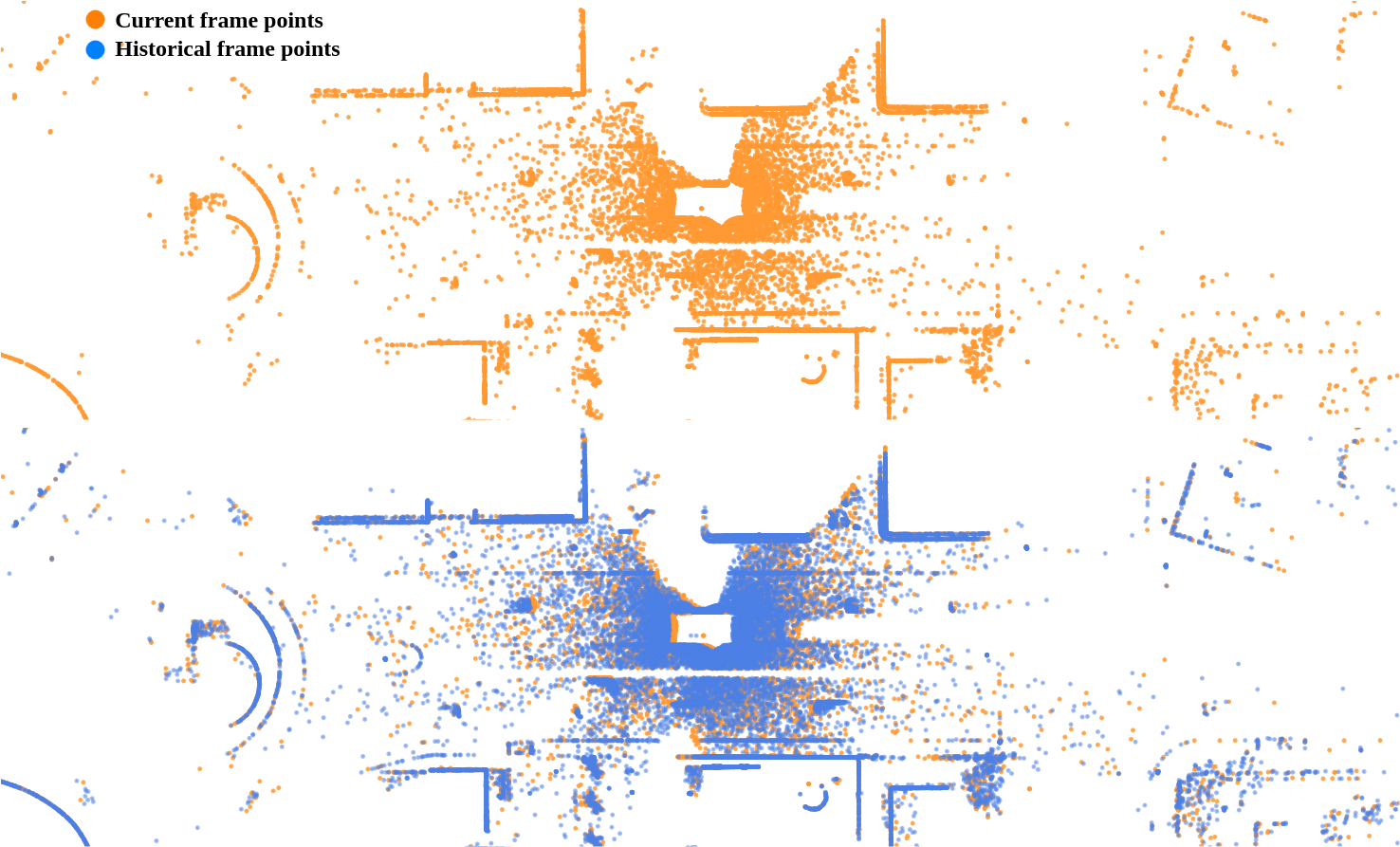}
    \caption{
    Comparison of point clouds before augmentation (top only with current points) and after augmentation (bottom with both current and compensated historical points). 
    With motion compensation, historical points are aligned to the current frame, resulting in a denser point cloud.
    }
    \label{fig:his_points}
\end{figure}

\subsection{Radar Aggregation with Historical Frames}
4D radar inherently suffers from point cloud sparsity, often leading to inaccurate and low-quality bounding box predictions~\cite{yang2020radarnetexploitingradarrobust}. 
To address this, we leverage multiple historical frames to densify radar point clouds. 

Each radar return provides spatial coordinates $(x,y,z)$ and a relative radial velocity $v_{\text{rel}}$ (Doppler velocity). 
Since Doppler encodes relative motion only along the sensor line-of-sight, we compute the ego-compensated velocity $v_{r}$ in the global frame as
\begin{equation}
  v_{r} = v_{\text{rel}} + (\mathbf v_{\text{ego}} \cdot \mathbf u),
\end{equation}
where $\mathbf v_{\text{ego}}$ is the ego-vehicle velocity and $\mathbf u=(u_x,u_y,u_z)$ is the unit vector from the sensor to the target. 
The resulting $v_{r}$ denotes the point’s absolute velocity along the line-of-sight. 
A point is classified as dynamic if the absolute value $|v_{r}| > \epsilon$, with $\epsilon$ a small threshold. 

In data augmentation, static points from historical frames remain unchanged since they do not move over time. 
In contrast, historical dynamic points are compensated along the line-of-sight using their ego-compensated velocity $v_{r}$ and the time gap ${\Delta}t$:
\begin{equation}
    (x',y',z') = (x,y,z) + v_{r} \cdot \Delta t \cdot \mathbf{u}.
\end{equation}

Subsequently, the compensated historical point clouds are transformed into the current frame’s coordinate system and concatenated with the current point cloud to obtain the final aggregated data. The aggregation result is presented in~\cref{fig:his_points}, while the clustered dynamic/static map is shown in the Point-Level Dynamic Map of~\cref{fig:mask}.

\begin{figure}[b!]
    \centering
    \includegraphics[width=0.9\linewidth]{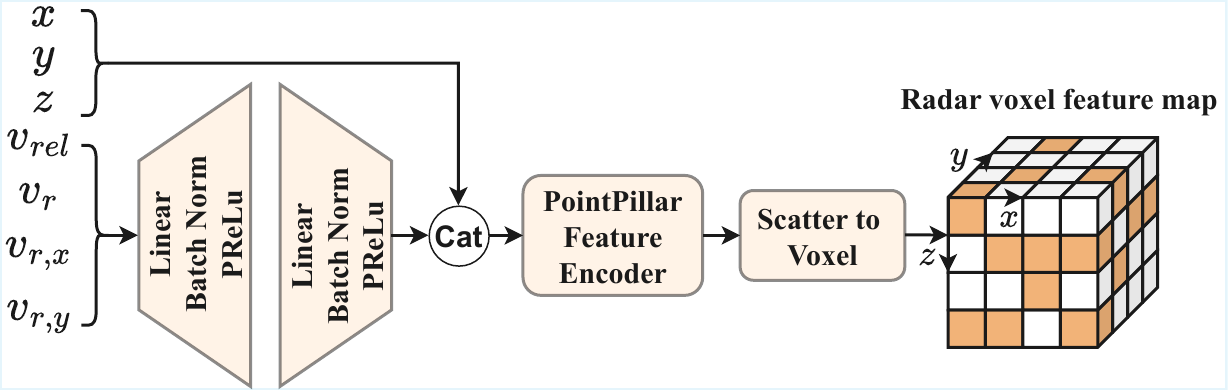}
    \caption{
        Overview of Doppler-aware motion encoding. 
        Radar returns provide Doppler velocities that are ego-compensated 
        and decomposed into planar components. These cues are embedded 
        by a velocity MLP and further used 
        to generate a voxel feature map.
    }
    \label{fig:vfe}
\end{figure}

\subsection{Velocity-Conditioned PillarVFE}
Following RadarPillars~\cite{radar_pillar}, we enrich radar point representations with velocity features to capture direction-aware motion cues beyond raw Doppler. 
The compensated absolute radial velocity $v_r$ provides stable motion estimates, while its Cartesian components $(v_{r,x}, v_{r,y})= (u_x, u_y) \cdot v_r$ explicitly encode ground-plane motion direction, making object dynamics more interpretable for the network.

As shown in \cref{fig:vfe}, we collect these descriptors into a velocity vector
\begin{equation}
  \mathbf v = (v_{\text{rel}}, v_r, v_{r,x}, v_{r,y}) \in \mathbb R^4,
\end{equation}
normalize it, and concatenate with spatial coordinates to form the radar feature
\begin{equation}
  \mathbf f = (x,y,z,\, \mathbf v) \in \mathbb R^7.
\end{equation}
A lightweight two-layer MLP with BatchNorm and PReLU refines the velocity embedding:
\begin{equation}
   \hat{\mathbf v} = f_\theta(\mathbf v).
\end{equation}
This design preserves Doppler sign, compensates scale differences, and enables the network to non-linearly reweight motion cues. The resulting embedding $\hat{\mathbf v}$ is combined with geometry and offsets before pillar max-pooling.

\subsection{Doppler Mask Generation}
\begin{figure*}[t!]
    \centering
    \includegraphics[width=1\linewidth]{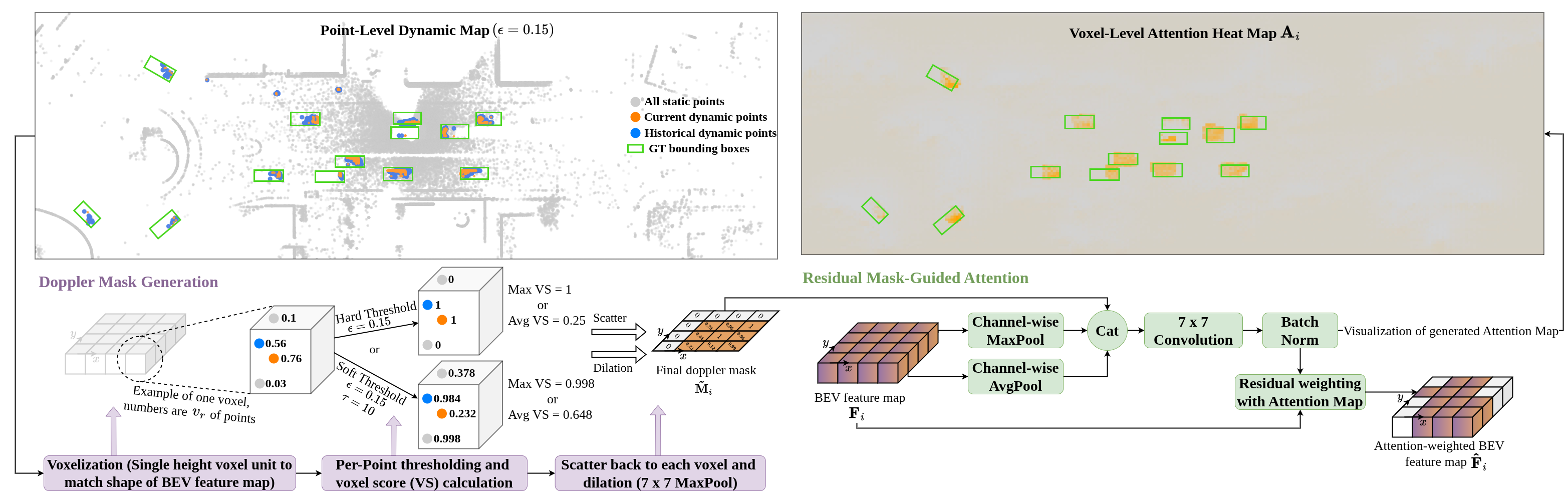}
    \caption{
    Overview of Doppler Mask Generation and Mask-Guided Spatial Attention.
    The Point-Level Dynamic Map illustrates how augmented historical data refine the representation of dynamic points, particularly in distant regions.
    The Voxel-Level Attention Heatmap demonstrates how the Doppler mask emphasizes moving objects and effectively guides spatial attention within the fusion backbone.}
    \label{fig:mask}
\end{figure*}

For each voxel, we compute a motion saliency score from the compensated velocity features. 
Depending on the configuration, we apply a soft score 
$\sigma(\tau (|v_r| - \epsilon))$, where $\sigma$ is the logistic function and $\tau$ 
controls steepness. Voxel scores are then reduced across all points within a voxel 
(by mean or max), yielding a per-voxel confidence value in $[0,1]$. 
These values are scattered onto the BEV grid to form a dense saliency map 
$\mathbf M$. To capture the spatial extent of objects 
and consolidate fragmented activations, we optionally apply morphological dilation 
(max-pooling), producing the final dynamic mask $\tilde{\mathbf M}$. An explicit visualization of this process is provided in~\cref{fig:mask}.

\subsection{Multi-Stage Mask-Guided Attention}
The dilated Doppler mask $\tilde{\mathbf M}$ is injected into the network at three stages, ensuring that motion cues are exploited at multiple abstraction levels:

\paragraph{Pre-Fusion Residual Gating.} BEV backbone features $\mathbf S$ are modulated before multi-agent fusion:
    \begin{equation}
      \mathbf{S} \leftarrow \mathbf{S} \odot (1 + \gamma_{\text{pre}} \tilde{\mathbf M}),
    \end{equation}
    with $\gamma_{\text{pre}} \ge 0$ learnable. This early gating primes the backbone to focus on dynamic regions from the outset.

\paragraph{Channel Gating.} A CBAM-style channel gate reweights feature channels using global average and max pooling followed by a shared MLP. While spatial gating highlights \emph{where} to focus, this stage emphasizes \emph{which} feature types (e.g., density, edges) are most informative.

\paragraph{Residual Mask-Guided Spatial Attention.} On the fused BEV $\mathbf F$, we concatenate average- and max-pooled maps with $\tilde{\mathbf M}$, apply a $7\times 7$ convolution and normalization, and obtain an attention map $\mathbf A$:
    \begin{equation}
        \text{\footnotesize $\mathbf{A} = \sigma\!\Big(\text{Norm}(\text{Conv}_{7\times 7}([\,\text{AvgPool}(\mathbf{F}),\, \text{MaxPool}(\mathbf{F}),\, \tilde{\mathbf M}\,]))\Big)$}.
    \end{equation}

 A residual weighting formulation preserves static context while emphasizing motion-relevant regions:
    \begin{equation}
        \hat{\mathbf{F}} = \mathbf{F} \odot \big(1 + \gamma \mathbf{A}\big), \qquad \gamma \ge 0.
    \end{equation}
This final stage aligns semantic content with Doppler-derived motion saliency. This process is illustrated in detail in~\cref{fig:mask}.

\subsection{Backbone Integration}
The above radar encoder, Doppler mask, and attention modules are designed as 
plug-and-play components. Their integration differs slightly between backbones:
\paragraph{BM2CP (radar-camera).} 
The LiDAR branch is replaced by a radar backbone. 
Pre-fusion gating is applied after radar-camera projection into BEV space, 
channel gating operates within the backbone, 
and residual spatial attention refines fused BEV features before collaborative detection.
\paragraph{Where2comm (LiDAR-radar).} 
Radar voxels are concatenated with LiDAR features, while a Doppler mask is computed in parallel. 
Pre-fusion gating modulates the combined BEV features before backbone processing and 
communication, channel gating follows feature compression, and spatial attention 
operates before the spatial confidence map is generated.
\section{Experiments}
\label{sec:evaluation}
We conducted a comprehensive evaluation of our radar-augmented collaborative perception framework. 
Our objectives are threefold: (i) assess model robustness under adverse weather, (ii) analyze the impact of different weather types on detection accuracy, and (iii) evaluate generalization from clear-weather training to adverse-weather conditions. 
In addition, we quantify the contribution of radar-camera fusion, LiDAR-radar fusion, and our mask-guided multi-stage Doppler attention mechanism.

\subsection{Sensor-Level Weather Augmentation} 
In CARLA~\cite{carla}, adverse weather primarily affects the camera stream, while LiDAR and radar remain idealized. LiDAR is simulated via geometric ray-casting without scattering, and radar lacks wave-particle interaction modeling. As a result, rain or fog has no effect on these modalities in synthetic datasets such as Adver-City~\cite{karvat2025advercityopensourcemultimodaldataset}, limiting realism.  

To address this, we augment LiDAR with the LISA framework~\cite{kilic2021lidar}, which models scattering-based degradation via Monte Carlo simulation. Rain is simulated using Mie scattering and the Marshall-Palmer raindrop distribution, producing attenuation, range-dependent point loss, and spurious near-range returns. Following~\cite{zhang2023perception}, we set rainfall intensities to $12\,\mathrm{mm/h}$ (light) and $40\,\mathrm{mm/h}$ (heavy). For fog, LISA applies homogeneous attenuation and gamma-distributed range noise, reducing effective LiDAR range. In compound conditions, rain and fog models are applied sequentially.  

Radar, by contrast, is minimally affected by weather. Studies~\cite{zhang2023perception} report only negligible impacts for radar in fog/rain, versus moderate-to-severe effects for LiDAR and cameras. Real-world datasets such as MAN TruckScenes~\cite{fent2024mantruckscenesmultimodaldataset} confirm this resilience, with radar detection stability across fog and rain. Given this robustness and the absence of validated radar degradation models, we omit radar augmentation to avoid introducing artifacts. Modeling radar degradation without validated physics-based models would be speculative and risk unfairly biasing the benchmark, whereas empirical studies confirm radar’s stability in fog and rain.

The resulting Adver-City-R dataset, therefore, includes physically grounded weather degradation for cameras and LiDAR, while radar remains intact as a robust modality. This provides a consistent and realistic benchmark for evaluating collaborative perception in adverse conditions, underscoring radar’s complementary role when LiDAR degrades, consistent with the physical sensor behaviors observed in operational environments.
\subsection{Training and Evaluation Strategy}
We consider two representative backbones: \textbf{BM2CP} (LiDAR-camera) and \textbf{Where2comm} (LiDAR-only).  
For each, we integrate 4D radar either as a replacement (BM2CP: radar-camera) or as a complement (Where2comm: LiDAR-radar), and evaluate both with and without our Doppler-guided attention. 

All models are trained on the \textbf{OPV2V-R} dataset (clear weather) and tested in a zero-shot manner on \textbf{Adver-City-R}, which includes soft rain, heavy rain, and fog. 

\subsection{Implementation Details}
All experiments were implemented in PyTorch and executed on a workstation with an NVIDIA GeForce RTX 4090 GPU. Each baseline model was trained using the official settings and hyperparameters provided in its original publication to ensure reproducibility. Both LiDAR and radar point clouds are voxelized with a horizontal resolution of $0.4 \times 0.4$~m and a vertical resolution of $4.0$~m. The perception range is set to $[-40, 40]$~m laterally, $[-4, 4]$~m vertically, and $[-140, 140]$~m longitudinally. Multi-agent cooperation is limited to a 70~m maximum communication range, with up to five connected vehicles in each scenario. Standard LiDAR geometric augmentations, such as rotation and translation, were not applied, since such transformations would distort Doppler velocity patterns and disrupt the spatial alignment required for collaboration~\cite{VoD}.

\subsection{Quantitative Results and Analysis}
Unless stated otherwise, we report \textit{Average Precision} (AP) at three IoU thresholds, $\{0.3,0.5,0.7\}$, following the OPV2V/Adver-City evaluation protocol.  For clarity, we always indicate whether the score refers to the \textit{Sparse} (S), \textit{Dense} (D), or \textit{Combined} (C) split. These splits correspond to low (S) and high (D) traffic densities in the simulation scenarios.

\subsubsection{Baseline under Clear Weather: OPV2V-R}\label{sec:opv2v_clear}
To establish a baseline under clear-weather conditions, we evaluate both BM2CP and Where2comm backbones on the OPV2V-R test split. As shown in~\cref{tab:opv2v_combined}, the LiDAR-based BM2CP achieves the highest accuracy among our variants, consistent with the strong spatial density of LiDAR point clouds. The radar-camera variant (BM2CP-RC) performs lower, but our extension with Doppler-guided masking and spatial attention (BM2CP-RCA) recovers a notable part of this gap, confirming that motion-aware masking strengthens radar perception even under ideal conditions. 

For completeness, we also report results for the Where2comm backbone and its radar-augmented variants. Unlike BM2CP, where radar substitutes LiDAR, in Where2comm radar complements LiDAR, providing an additional source of geometric information. Including these results demonstrates that our Doppler-guided masking mechanism is not tied to a single backbone design but can be integrated across different collaborative perception frameworks, yielding consistent benefits.

\begin{table}[h!]
\footnotesize
\caption{Performance on OPV2V-R (clear weather). AP is reported for three IoU thresholds: 0.3, 0.5, and 0.7. RC (radar-camera), RCA (RC + mask-guided spatial attention), LR (LiDAR-radar), LRA (LR + mask-guided spatial attention).}
\label{tab:opv2v_combined}
\centering
\begin{tabular}{@{}lccc@{}}
\toprule
\textbf{Model Variant} & \textbf{AP@0.3} & \textbf{AP@0.5} & \textbf{AP@0.7} \\
\midrule
BM2CP~\cite{zhao2023bm2cp} (LiDAR-camera) & \textbf{85.39} & \textbf{83.65} & \textbf{63.25}\\
BM2CP-RC & 78.19 & 75.59 & 55.77  \\
BM2CP-RCA & 82.89 & 77.95 & 58.80  \\
\midrule
Where2comm~\cite{hu2022where2comm} (LiDAR) & 87.10 & 86.00 & 73.50  \\
Where2comm-LR & 87.49 & 86.21 & \textbf{76.50} \\
Where2comm-LRA & \textbf{91.00} & \textbf{90.00} & 75.00 \\
\bottomrule
\end{tabular}
\end{table}

\subsubsection{Zero-shot generalization on Adver-City-R}\label{sec:zero_shot}
~\cref{tab:advercity_combined} reports zero-shot performance across adverse weather conditions. Note that the \textit{Clear} baseline in the table is evaluated on the Adver-City-R dataset. The lower absolute AP compared to \cref{tab:opv2v_combined} is attributed to the increased complexity in the scenarios~\cite{yazgan2025slimcomm}. Within the BM2CP family, the LiDAR-camera (LC) baseline remains superior in clear weather and rain, but performance collapses in fog (32.97 AP@0.5) due to LiDAR backscatter. Transitioning to radar-camera (RC) restores basic visibility, though our Doppler-guided RCA is the decisive factor for high precision, achieving the best overall performance in sparse fog (65.12 AP) and fog+rain (65.00 AP).

Notably, RCA surpasses the LiDAR-radar LRA model in these sparse settings, suggesting that a refined radar signal with intelligent attention can be more reliable than fusion hindered by severely degraded LiDAR features. For the Where2comm family, the LiDAR-only model drops sharply in fog ($\approx$36.00 AP); while the LRA variant dominates dense splits, the performance gap between LR and LRA confirms that Doppler-based attention is essential to align motion cues with spatial context. These results establish radar-augmented setups as competitive alternatives to tri-modal fusion in extreme conditions, proving that our attention mechanism is key to unlocking radar's full potential.
\subsection{Comparison with State-of-the-Art Methods}
\begin{table}[h!]
\centering
\footnotesize
\caption{Comparison with State-of-the-Art methods on Adver-City-R (Combined split). 
We report AP@0.5 and AP@0.7 for LiDAR-only, radar-camera, and LiDAR-radar baselines. 
Our Where2comm-LRA achieves the highest overall robustness.}
\label{tab:sota_comparison}
\begin{tabular}{lccc}
\toprule
\textbf{Method} & \textbf{Modality} & \textbf{AP@0.5} & \textbf{AP@0.7} \\
\midrule
Scope~\cite{yang2023spatio} & LiDAR & 19.60 & 14.40 \\
AttFuse~\cite{xu2022opencood} & LiDAR & 24.00 & 16.00 \\
S-AdaFusion~\cite{qiao2023adaptive} & LiDAR & 38.30 & 30.30 \\
SlimComm~\cite{yazgan2025slimcomm} & LiDAR-radar & 58.00 & 50.74\\
BM2CP-RCA & radar-camera & 55.36 & 40.10 \\
AttFuse w/MDD~\cite{huang2024v2x} & LiDAR-radar & 55.57 & 47.04 \\
Where2comm-LRA & LiDAR-radar & \textbf{64.70} & \textbf{51.12} \\
\bottomrule
\end{tabular}
\end{table}
\begin{table*}[t!]
\centering
\caption{Zero-shot generalization on Adver-City-R (AP@0.5, \%). 
Scores are reported for \textit{Sparse} (S) and \textit{Dense} (D) splits. 
Abbreviations: BM2CP (LiDAR-camera), RC (radar-camera), RCA (RC + mask-guided spatial attention), 
Where2comm (LiDAR-only), LR (LiDAR-radar), LRA (LR + mask-guided spatial attention). 
\textbf{Bold} = best, \underline{underline} = second-best; 
$^{\dagger}$ = best within BM2CP family; $^{\ddagger}$ = best within Where2comm family; 
$^{\star}$ = best within the respective S or D split across all models.}
\label{tab:advercity_combined}
\resizebox{\textwidth}{!}{%
\begin{tabular}{lcccccc}
\toprule
& BM2CP~\cite{zhao2023bm2cp} & BM2CP-RC & BM2CP-RCA & Where2comm~\cite{hu2022where2comm} & Where2comm-LR & Where2comm-LRA \\
\textbf{Weather} & \textbf{S/D} & \textbf{S/D} & \textbf{S/D} & \textbf{S/D} & \textbf{S/D} & \textbf{S/D} \\
\midrule
\midrule
Clear       
& 67.37$^{\dagger}$ / 61.97$^{\dagger}$  
& 60.64 / 47.18  
& 64.81 / 54.50  
& 69.54 / 69.15  
& \underline{72.51} / \underline{70.38}  
& \textbf{74.79}$^{\ddagger\star}$ / \textbf{70.91}$^{\ddagger\star}$ \\
Soft Rain   
& 64.39$^{\dagger}$ / 59.14$^{\dagger}$  
& 60.12 / 46.33  
& 63.92 / 53.20  
& 65.47 / 63.31  
& \underline{71.63} / \underline{67.00}  
& \textbf{73.81}$^{\ddagger\star}$ / \textbf{67.97}$^{\ddagger\star}$ \\
Heavy Rain  
& 65.27$^{\dagger}$ / 61.06$^{\dagger}$  
& 59.11 / 46.49  
& 63.91 / 54.39  
& 63.51 / 61.54  
& \underline{72.23} / \underline{64.97}  
& \textbf{75.20}$^{\ddagger\star}$ / \textbf{68.89}$^{\ddagger\star}$ \\
Fog         
& 32.97 / 34.32  
& 60.84 / 45.55  
& \textbf{65.12}$^{\dagger\star}$ / \underline{53.37}$^{\dagger}$  
& 36.00 / 35.77  
& 63.07 / 53.11  
& \underline{64.93}$^{\ddagger}$ / \textbf{56.80}$^{\ddagger\star}$ \\
Fog+Rain    
& 28.05 / 29.10  
& 59.72 / 46.01  
& \textbf{65.00}$^{\dagger\star}$ / \underline{52.03}$^{\dagger}$  
& 33.60 / 31.34  
& 60.50 / 51.61  
& \underline{63.12}$^{\ddagger}$ / \textbf{54.00}$^{\ddagger\star}$ \\
\bottomrule
\end{tabular}%
}
\end{table*}
To contextualize our results, ~\cref{tab:sota_comparison} compares our radar-augmented models with representative LiDAR-based cooperative perception baselines, including Scope~\cite{yang2023spatio}, \mbox{AttFuse}~\cite{xu2022opencood}, and S-AdaFusion~\cite{qiao2023adaptive}, as well as LiDAR-radar baselines such as AttFuse w/MDD~\cite{huang2024v2x} and SlimComm~\cite{yazgan2025slimcomm} on the Adver-City-R benchmark (Combined split). Crucially, while SlimComm inherently supports $360^{\circ}$ perception, we retrained and evaluated AttFuse w/MDD on Adver-City-R with the same $360^{\circ}$ coverage and continuous Doppler velocity data used by our models. The LiDAR-only methods achieve moderate accuracy but degrade substantially in adverse weather. By contrast, our radar-based variants close this gap, with RCA already outperforming all LiDAR-only baselines, and our LiDAR-radar LRA model further exceeding AttFuse w/MDD while delivering the most robust overall results among the compared methods. These results highlight radar’s unique role in enabling reliable cooperative perception beyond LiDAR-only pipelines and show that our mask-guided Doppler attention consistently improves performance under adverse weather conditions. Furthermore, our Where2comm-LRA demonstrates superior computational efficiency with an inference time of 58 ms, nearly twice as fast as the AttFuse w/MDD baseline (108 ms). This shows that our Doppler-guided attention is a lightweight alternative to high-overhead diffusion-based denoising.
\subsection{Robustness to Spatial and Temporal Misalignment}
The framework is evaluated under simulated spatial and temporal misalignments on the Adver-City-R combined fog and heavy rain split based on protocol in AttFuse w/MDD \cite{huang2024v2x}. Proposed RCA and LRA models are compared against SlimComm \cite{yazgan2025slimcomm} and AttFuse w/MDD \cite{huang2024v2x}.

\begin{figure}[t]
\centering
\includegraphics[width=\columnwidth]{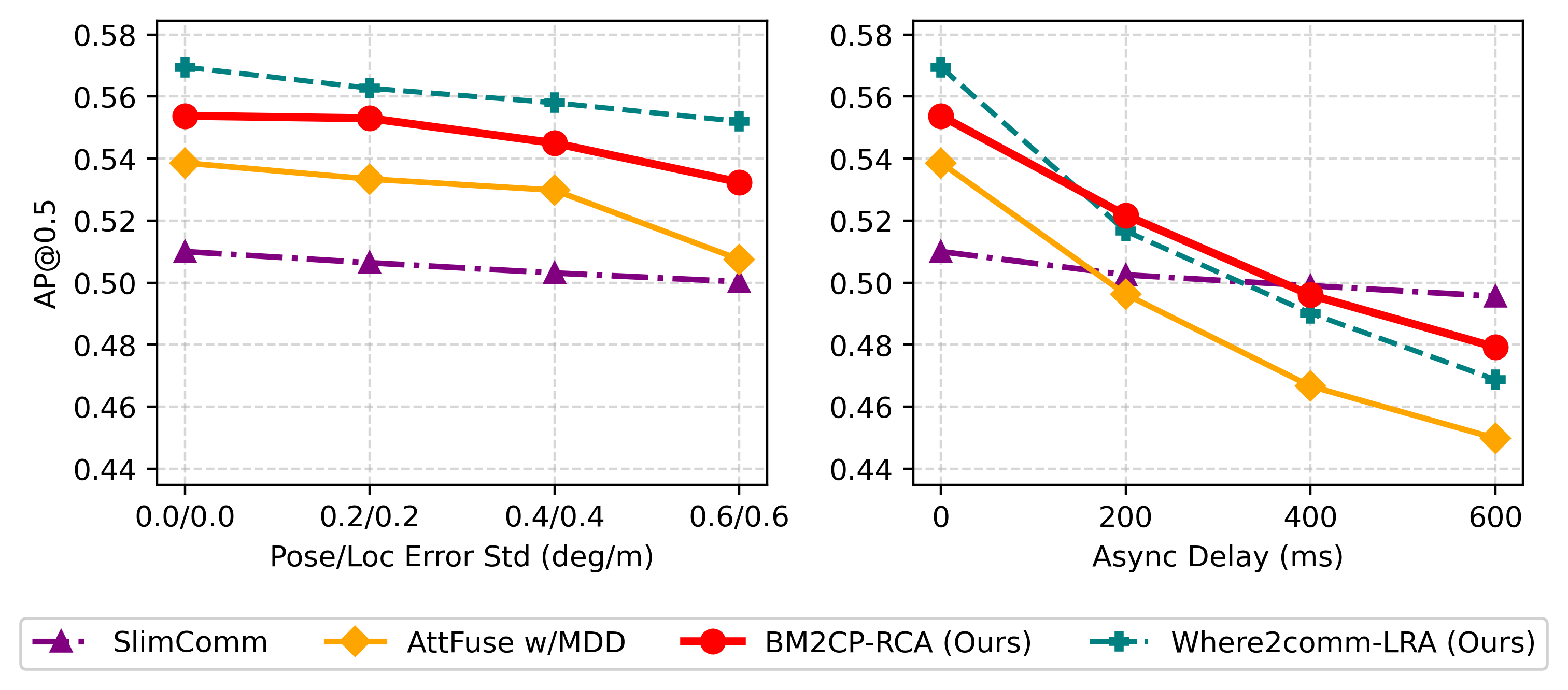}
\caption{Robustness analysis on Adver-City-R (Combined Fog and Heavy Rain). (Left) Impact of increasing pose and localization error. (Right) Impact of asynchronous communication delay.}
\label{fig:pose_robustness}
\end{figure}

As shown in~\cref{fig:pose_robustness} (Left), all models experience performance degradation as pose and localization error increase to 0.6$^{\circ}$/0.6 m. Where2comm-LRA maintains the highest absolute AP@0.5 throughout the entire error range. While SlimComm exhibits high relative stability indicated by a flatter slope, it consistently performs at a lower baseline. In contrast, BM2CP-RCA provides a superior balance, maintaining a significantly higher AP than both SlimComm and AttFuse w/MDD while exhibiting comparable stability to increasing pose noise.

The impact of communication latency, illustrated in~\cref{fig:pose_robustness} (Right), reveals a critical shift in model ranking. While Where2comm-LRA is superior under ideal synchronization (0 ms), it is highly sensitive to temporal shifts. SlimComm proves to be the most resilient model with a relatively stable performance curve, likely due to its sparse query mechanism, which is less dependent on perfect temporal feature alignment. 
However, BM2CP-RCA emerges as the most effective solution for high-latency environments. It crosses over both the LRA and AttFuse w/MDD models at the 200 ms mark, maintaining the highest AP@0.5 for all delays between 200 ms and 600 ms. This suggests that the Doppler-guided attention in RCA effectively mitigates ghosting artifacts by using instantaneous velocity cues to anchor features, even when the received messages are significantly delayed.
\subsection{Ablation Study}
\label{sec:ablation}
To better understand the contribution of each component in our radar-camera pipeline, we performed an ablation analysis. 
As shown in~\cref{tab:ablation_acc_three}, mask-guided spatial attention already improves over plain RC, velocity conditioning yields an additional gain, and temporal aggregation provides the strongest boost. 
This confirms that Doppler-based history compensation is crucial for robustness under adverse weather.
\begin{table}[h!]
\footnotesize
\centering
\caption{The table shows the contribution of each module: \textbf{Vel} (velocity-conditioned PillarVFE), \textbf{A} (mask-guided spatial attention), and \textbf{Temp} (Doppler-compensated multi-frame aggregation).}
\label{tab:ablation_acc_three}
\begin{tabular}{lcccc}
\toprule
\textbf{Variant} & \textbf{Vel} & \textbf{A} & \textbf{Temp} & \textbf{AP@0.5} \\
\midrule
LC (LiDAR-camera baseline) & & & & 48.86 \\
RC (radar-camera) & & & & 47.34 \\
RC + A & & \checkmark & & 52.80 \\
RC + Vel + A & \checkmark & \checkmark & & 53.15 \\
\textbf{RCA (Full Model)} & \checkmark & \checkmark & \checkmark & \textbf{55.36} \\
\bottomrule
\end{tabular}
\end{table}

\subsection{Real-World Single-Vehicle Evaluation}
\label{sec:real_world}

\noindent\textbf{Motivation.}
Adver-City-R and OPV2V-R quantify robustness under controlled, cooperative settings with weather degradation. To test real-world applicability beyond simulation, we additionally evaluate our radar encoding and Doppler-guided modules on the real MAN TruckScenes dataset~\cite{fent2024mantruckscenesmultimodaldataset}. Since MAN TruckScenes does not include cooperative labels, we use a single-vehicle PointPillars-based~\cite{ppillars} backbone for this study.
\smallskip

\noindent\textbf{Setup.}
MAN TruckScenes follows a nuScenes-like structure. We use all six radar and six LiDAR sensors. Ground truth is limited to the ``car'' class and includes only objects hit by at least one LiDAR or radar point. We select two subsets of the dataset (approximately 3960 frames), filter scenes to highway, city, residential, and rural areas, and split 80/10/10 with a fixed seed. Following the simulation protocol, all models are trained on clear-weather data and evaluated zero-shot on the foggy subset. Perception range and PointPillars settings are aligned with nuScenes. For the evaluation, we used mAP-based results to align with~\cref{sec:opv2v_clear,sec:zero_shot}. All experiments were conducted on an NVIDIA RTX~2080~Ti GPU. Latency is reported as the mean model-only forward pass latency, averaged over 100 iterations.
\begin{table}[h!]
\footnotesize
\centering
\caption{MAN TruckScenes (single-vehicle, car class only) with a PointPillars-based (PP) backbone~\cite{ppillars}.
The table shows the contribution of each module: \textbf{Vel} (velocity-conditioned PillarVFE), \textbf{A} (mask-guided spatial attention), and \textbf{Temp} (Doppler-compensated multi-frame aggregation).
Results are AP@0.5 for Clear/Overcast and Foggy subsets.}
\label{tab:truckscenes}
\resizebox{\columnwidth}{!}{%
\begin{tabular}{lccc}
\toprule
\textbf{Model Variant} & \textbf{Clear/Overcast} & \textbf{Foggy} & \textbf{Latency (ms)} \\
\midrule
PP-L (LiDAR-only)                & 17.70 & 25.82 & 14.0 \\
PP-LR (LiDAR-radar)              & 17.40 & 24.52 & 14.7 \\
PP-LR + A                        & 20.15 & 30.98 & 16.9 \\
PP-LR + A + Vel                  & 20.15 & 31.92 & 17.0 \\
PP-LR + A + Vel + Temp           & \textbf{21.20} & \textbf{33.80} & 17.1 \\
\bottomrule
\end{tabular}%
}
\end{table}

\noindent\textbf{Findings.}
~\cref{tab:truckscenes} shows that naive LiDAR-radar fusion (PP-LR) does not improve over LiDAR-only (PP-L), and in fact performs slightly worse in both Clear/Overcast and Foggy conditions. This indicates that raw radar features alone are not directly beneficial in the real-world dataset. In contrast, adding our mask-guided spatial attention (+A) yields consistent gains (+2.45 AP on Clear/Overcast, +5.16 AP on Foggy) while increasing latency by only $\sim$2\, ms. Velocity conditioning (+Vel) further improves robustness, reaching with just $\sim$3\,ms overhead compared to the LiDAR-only baseline. Temporal aggregation (+Temp) provides the highest overall accuracy (21.20 AP on Clear/Overcast, 33.80 AP on Foggy), resulting in a total improvement of +8.0 points over the LiDAR-only baseline in Foggy conditions while adding only $\sim$3\, ms latency. 

Note that, unlike Adver-City-R, where weather is applied to identical scenarios, MAN TruckScenes does not control for scene type. The Foggy split contains only highway scenes, which have fewer occlusions and targets, leading to higher LiDAR AP than in Clear/Overcast. This reflects dataset composition rather than LiDAR robustness to fog; the relevant finding is the consistent gain from our Doppler-guided modules.

\section{Limitations and Future Work}
\label{sec:future_works}
While our method shows strong robustness under adverse weather, several limitations remain. Most experiments are conducted on radar-augmented simulation benchmarks, and the real-world validation on MAN TruckScenes is limited to a single-vehicle setup with car-only annotations. In addition, we do not explicitly model radar-specific artifacts such as multipath reflections or sensor-dependent noise. Future research will explore backbone-aware attention and advanced temporal aggregation~\cite{doppdrive2025} to better handle real-world radar sparsity and noise. We also plan to extend our cooperative benchmarks to more challenging conditions, including snow, roadside infrastructure, and Vulnerable Road Users (VRUs), and to study radar motion cues for downstream tasks such as collaborative tracking and trajectory prediction.
\section{Conclusion}
\label{sec:conclusion}
We addressed adverse-weather cooperative perception by integrating 4D radar into BM2CP and Where2comm. Our Doppler-guided spatial attention improves robustness to degraded LiDAR and camera observations. Experiments on OPV2V-R and Adver-City-R show strong gains under fog and rain, and MAN TruckScenes confirms transfer beyond simulation. The proposed LRA model delivers robust and efficient all-weather cooperative perception. 
\paragraph{Acknowledgements.}This paper was created in the Country 2 City - Bridge project of the German Center for Future Mobility, which is funded by the German Federal Ministry for Digital and Transport.
{
    \small
    \bibliographystyle{ieeenat_fullname}
    \bibliography{main}

@String(CVPR= {IEEE Conf. Comput. Vis. Pattern Recog.})

@String(ICCV= {Int. Conf. Comput. Vis.})

@String(AAAI = {AAAI})

@String(CVPR  = {CVPR})

@String(ICCV  = {ICCV})

@article{yazgan2024real,
  title={Real-World Problems in Collaborative Perception: A Categorized Review of Intermediate Fusion Methods},
  author={Yazgan, Melih and Graf, Thomas and Liu, Min and Fleck, Tobias and Z{\"o}llner, J Marius},
  journal={IEEE IV},
  year={2024}
}

@inproceedings{zhao2023bm2cp,
  title={BM2CP: Efficient Collaborative Perception with LiDAR-Camera Modalities},
  author={Zhao, Binyu and ZHANG, Wei and Zou, Zhaonian},
  booktitle={Conference on Robot Learning},
  pages={1022--1035},
  year={2023},
  organization={PMLR}
}

@article{hu2022where2comm,
  title={Where2comm: Communication-efficient collaborative perception via spatial confidence maps},
  author={Hu, Yue and Fang, Shaoheng and Lei, Zixing and Zhong, Yiqi and Chen, Siheng},
  journal={Advances in neural information processing systems},
  volume={35},
  pages={4874--4886},
  year={2022}
}

@inproceedings{xu2022opencood,
  title={Opv2v: An open benchmark dataset and fusion pipeline for perception with vehicle-to-vehicle communication},
  author={Xu, Runsheng and Xiang, Hao and Xia, Xin and Han, Xu and Li, Jinlong and Ma, Jiaqi},
  booktitle={2022 International Conference on Robotics and Automation (ICRA)},
  pages={2583--2589},
  year={2022},
  organization={IEEE}
}

@inproceedings{carla,
  title = {{CARLA}: {An} Open Urban Driving Simulator},
  author = {Alexey Dosovitskiy and German Ros and Felipe Codevilla and Antonio Lopez and Vladlen Koltun},
  booktitle = {Proceedings of the 1st Annual Conference on Robot Learning},
  pages = {1--16},
  year = {2017}
}

@INPROCEEDINGS{YazganCoPeData,
  author={Yazgan, Melih and Akkanapragada, Mythra Varun and Marius Zöllner, J.},
  booktitle={2024 IEEE Intelligent Vehicles Symposium (IV)}, 
  title={Collaborative Perception Datasets in Autonomous Driving: A Survey}, 
  year={2024},
  volume={},
  number={},
  pages={2269-2276},
  doi={10.1109/IV55156.2024.10588870}}

@article{Liu_2025,
  title={Autonomous vehicles: A critical review (2004-2024) and a vision for the future},
  author={Liu, Henry and Cao, Zhong and Yan, Xintao and Feng, Shuo and Lu, Qiujing},
  journal={Authorea Preprints},
  year={2025},
  publisher={Authorea}
}

@article{wu2025foundationmodelsautonomousdriving,
  title={Foundation Models for Autonomous Driving Systems: An Initial Roadmap},
  author={Wu, Xiongfei and Cheng, Mingfei and Ren, Xiaoning and Hu, Qiang and Chen, Jianlang and Huang, Yuheng and Cordy, Maxime and Zhang, Yao and Xie, Xiaofei and Ma, Lei and others},
  journal={ACM Transactions on Software Engineering and Methodology},
  year={2026},
  publisher={ACM New York, NY}
}

@misc{peng20254dmmwaveradarsensing,
	title        = {4D mmWave Radar for Sensing Enhancement in Adverse Environments: Advances and Challenges},
	author       = {Xiangyuan Peng and Miao Tang and Huawei Sun and Kay Bierzynski and Lorenzo Servadei and Robert Wille},
	year         = 2025,
	url          = {https://arxiv.org/abs/2503.24091},
	eprint       = {2503.24091},
	archiveprefix = {arXiv},
	primaryclass = {cs.CV}
}

@misc{huang2024v2xcooperativeperceptionautonomous,
	title        = {V2X Cooperative Perception for Autonomous Driving: Recent Advances and Challenges},
	author       = {Tao Huang and Jianan Liu and Xi Zhou and Dinh C. Nguyen and Mostafa Rahimi Azghadi and Yuxuan Xia and Qing-Long Han and Sumei Sun},
	year         = 2024,
	url          = {https://arxiv.org/abs/2310.03525},
	eprint       = {2310.03525},
	archiveprefix = {arXiv},
	primaryclass = {cs.CV}
}

@article{liu2024surveyautonomousdrivingdatasets,
  title={A survey on autonomous driving datasets: Statistics, annotation quality, and a future outlook},
  author={Liu, Mingyu and Yurtsever, Ekim and Fossaert, Jonathan and Zhou, Xingcheng and Zimmer, Walter and Cui, Yuning and Zagar, Bare Luka and Knoll, Alois C},
  journal={IEEE Transactions on Intelligent Vehicles},
  year={2024},
  publisher={IEEE}
}

@misc{sun2020scalabilityperceptionautonomousdriving,
	title        = {Scalability in Perception for Autonomous Driving: Waymo Open Dataset},
	author       = {Pei Sun and Henrik Kretzschmar and Xerxes Dotiwalla and Aurelien Chouard and Vijaysai Patnaik and Paul Tsui and James Guo and Yin Zhou and Yuning Chai and Benjamin Caine and Vijay Vasudevan and Wei Han and Jiquan Ngiam and Hang Zhao and Aleksei Timofeev and Scott Ettinger and Maxim Krivokon and Amy Gao and Aditya Joshi and Sheng Zhao and Shuyang Cheng and Yu Zhang and Jonathon Shlens and Zhifeng Chen and Dragomir Anguelov},
	year         = 2020,
	url          = {https://arxiv.org/abs/1912.04838},
	eprint       = {1912.04838},
	archiveprefix = {arXiv},
	primaryclass = {cs.CV}
}

@article{wang2020fusion,
	title        = {Multi-Sensor Fusion in Automated Driving: A Survey},
	author       = {Wang, Zhangjing and Wu, Yu and Niu, Qingqing},
	year         = 2020,
	month        = {},
	journal      = {IEEE Access},
	volume       = 8,
	number       = {},
	pages        = {2847--2868},
	doi          = {10.1109/ACCESS.2019.2962554},
	issn         = {2169-3536}
}

@article{wevj2025,
	title        = {A Review of Environmental Perception Technology Based on Multi-Sensor Information Fusion in Autonomous Driving},
	author       = {Yang, Boquan and Li, Jixiong and Zeng, Ting},
	year         = 2025,
	journal      = {World Electric Vehicle Journal},
	volume       = 16,
	number       = 1,
	doi          = {10.3390/wevj16010020},
	issn         = {2032-6653},
	url          = {https://www.mdpi.com/2032-6653/16/1/20},
	article-number = 20
}

@INPROCEEDINGS{huang2024v2x,
  author={Huang, Xun and Wang, Jinlong and Xia, Qiming and Chen, Siheng and Yang, Bisheng and Li, Xin and Wang, Cheng and Wen, Chenglu},
  booktitle={2025 IEEE/CVF Conference on Computer Vision and Pattern Recognition (CVPR)}, 
  title={V2X-R: Cooperative LiDAR-4D Radar Fusion with Denoising Diffusion for 3D Object Detection}, 
  year={2025},
  volume={},
  number={},
  pages={27390-27400},
  doi={10.1109/CVPR52734.2025.02551}}

@article{Yao_2024,
	title        = {Radar-Camera Fusion for Object Detection and Semantic Segmentation in Autonomous Driving: A Comprehensive Review},
	author       = {Yao, Shanliang and Guan, Runwei and Huang, Xiaoyu and Li, Zhuoxiao and Sha, Xiangyu and Yue, Yong and Lim, Eng Gee and Seo, Hyungjoon and Man, Ka Lok and Zhu, Xiaohui and Yue, Yutao},
	year         = 2024,
	month        = jan,
	journal      = {IEEE Transactions on Intelligent Vehicles},
	publisher    = {Institute of Electrical and Electronics Engineers (IEEE)},
	volume       = 9,
	number       = 1,
	pages        = {2094–2128},
	doi          = {10.1109/tiv.2023.3307157},
	issn         = {2379-8858},
	url          = {http://dx.doi.org/10.1109/TIV.2023.3307157}
}

@misc{karvat2025advercityopensourcemultimodaldataset,
	title        = {Adver-City: Open-Source Multi-Modal Dataset for Collaborative Perception Under Adverse Weather Conditions},
	author       = {Mateus Karvat and Sidney Givigi},
	year         = 2025,
	url          = {https://arxiv.org/abs/2410.06380},
	eprint       = {2410.06380},
	archiveprefix = {arXiv},
	primaryclass = {cs.CV}
}

@inproceedings{dair-v2x,
  title={Dair-v2x: A large-scale dataset for vehicle-infrastructure cooperative 3d object detection},
  author={Yu, Haibao and Luo, Yizhen and Shu, Mao and Huo, Yiyi and Yang, Zebang and Shi, Yifeng and Guo, Zhenglong and Li, Hanyu and Hu, Xing and Yuan, Jirui and Nie, Zaiqing},
  booktitle={Proceedings of the IEEE/CVF Conference on Computer Vision and Pattern Recognition},
  pages={21361--21370},
  year={2022}
}

@InProceedings{Zimmer_2024_CVPR,
    author    = {Zimmer, Walter and Wardana, Gerhard Arya and Sritharan, Suren and Zhou, Xingcheng and Song, Rui and Knoll, Alois C.},
    title     = {TUMTraf V2X Cooperative Perception Dataset},
    booktitle = {Proceedings of the IEEE/CVF Conference on Computer Vision and Pattern Recognition (CVPR)},
    month     = {June},
    year      = {2024},
    pages     = {22668-22677}
}

@article{wang2025collaborativeperceptiondatasetsautonomous,
  title={Collaborative perception datasets for autonomous driving: A review},
  author={Wang, Naibang and Shang, Deyong and Gong, Yan and Hu, Xiaoxi and Song, Ziying and Yang, Lei and Huang, Yuhan and Wang, Xiaoyu and Lu, Jianli},
  journal={IEEE Sensors Journal},
  year={2025},
  publisher={IEEE}
}

@article{kilic2021lidar,
  title={Lidar light scattering augmentation (lisa): Physics-based simulation of adverse weather conditions for 3d object detection},
  author={Kilic, Velat and Hegde, Deepti and Sindagi, Vishwanath and Cooper, A Brinton and Foster, Mark A and Patel, Vishal M},
  journal={arXiv preprint arXiv:2107.07004},
  year={2021}
}

@inproceedings{ppillars,
	title        = {PointPillars: Fast Encoders for Object Detection From Point Clouds},
	author       = {Lang, Alex H. and Vora, Sourabh and Caesar, Holger and Zhou, Lubing and Yang, Jiong and Beijbom, Oscar},
	year         = 2019,
	month        = {June},
	booktitle    = {Proceedings of the IEEE/CVF Conference on Computer Vision and Pattern Recognition (CVPR)}
}

@inproceedings{qiao2023adaptive,
  title={Adaptive feature fusion for cooperative perception using lidar point clouds},
  author={Qiao, Donghao and Zulkernine, Farhana},
  booktitle={Proceedings of the IEEE/CVF winter conference on applications of computer vision},
  pages={1186--1195},
  year={2023}
}

@Inproceedings{yang2023spatio,
   author    = {Yang, Kun and Yang, Dingkang and Zhang, Jingyu and Li, Mingcheng and Liu, Yang and Liu, Jing and Wang, Hanqi and Sun, Peng and Song, Liang},
   title     = {Spatio-Temporal Domain Awareness for Multi-Agent Collaborative Perception},
   booktitle = {Proceedings of the IEEE/CVF International Conference on Computer Vision (ICCV)},
   month     = {October},
   year      = {2023},
   pages     = {23383-23392}
}

@INPROCEEDINGS{radar_pillar,
  author={Musiat, Alexander and Reichardt, Laurenz and Schulze, Michael and Wasenmüller, Oliver},
  booktitle={2024 IEEE 27th International Conference on Intelligent Transportation Systems (ITSC)}, 
  title={RadarPillars: Efficient Object Detection from 4D Radar Point Clouds}, 
  year={2024},
  volume={},
  number={},
  pages={1656-1663},
  doi={10.1109/ITSC58415.2024.10919920}}

@article{zhang2023perception,
  title={Perception and sensing for autonomous vehicles under adverse weather conditions: A survey},
  author={Zhang, Yuxiao and Carballo, Alexander and Yang, Hanting and Takeda, Kazuya},
  journal={ISPRS Journal of Photogrammetry and Remote Sensing},
  volume={196},
  pages={146--177},
  year={2023},
  publisher={Elsevier}
}

@INPROCEEDINGS{teufel2023weather,
  author={Teufel, Sven and Volk, Georg and Von Bernuth, Alexander and Bringmann, Oliver},
  booktitle={2022 IEEE 95th Vehicular Technology Conference: (VTC2022-Spring)}, 
  title={Simulating Realistic Rain, Snow, and Fog Variations For Comprehensive Performance Characterization of LiDAR Perception}, 
  year={2022},
  volume={},
  number={},
  pages={1-7},
  doi={10.1109/VTC2022-Spring54318.2022.9860868}}

@INPROCEEDINGS{tsakmakopoulou2023v2v,
  author={Tsakmakopoulou, Dimitra and Moustakas, Konstantinos},
  booktitle={2024 IEEE/RSJ International Conference on Intelligent Robots and Systems (IROS)}, 
  title={Perception for Connected Autonomous Vehicles under Adverse Weather Conditions}, 
  year={2024},
  volume={},
  number={},
  pages={3161-3166},
  doi={10.1109/IROS58592.2024.10801295}}

@ARTICLE{jiang2023weather,
  author={Jiang, Ping and Deng, Xiaoheng and Wu, Weishang and Lin, Lixin and Chen, Xuechen and Chen, Chen and Wan, Shaohua},
  journal={IEEE Transactions on Intelligent Transportation Systems}, 
  title={Weather-Aware Collaborative Perception With Uncertainty Reduction}, 
  year={2024},
  volume={25},
  number={12},
  pages={20059-20072},
  doi={10.1109/TITS.2024.3479720}}

@misc{li2025dg,
      title={V2X-DGW: Domain Generalization for Multi-agent Perception under Adverse Weather Conditions}, 
      author={Baolu Li and Jinlong Li and Xinyu Liu and Runsheng Xu and Zhengzhong Tu and Jiacheng Guo and Xiaopeng Li and Hongkai Yu},
      year={2025},
      eprint={2403.11371},
      archivePrefix={arXiv},
      primaryClass={cs.CV},
      url={https://arxiv.org/abs/2403.11371}, 
}

@Article{wang2024sensors,
AUTHOR = {Wang, Jizhao and Wu, Zhizhou and Liang, Yunyi and Tang, Jinjun and Chen, Huimiao},
TITLE = {Perception Methods for Adverse Weather Based on Vehicle Infrastructure Cooperation System: A Review},
JOURNAL = {Sensors},
VOLUME = {24},
YEAR = {2024},
NUMBER = {2},
ARTICLE-NUMBER = {374},
URL = {https://www.mdpi.com/1424-8220/24/2/374},
PubMedID = {38257469},
ISSN = {1424-8220},
DOI = {10.3390/s24020374}
}

@inproceedings{yazgan2025slimcomm,
  title={SlimComm: Doppler-Guided Sparse Queries for Bandwidth-Efficient Cooperative 3-D Perception},
  author={Yazgan, Melih and Wu, Qiyuan and Hamdard, Iramm and Li, Shiqi and Zoellner, J Marius},
  booktitle={Proceedings of the IEEE/CVF International Conference on Computer Vision},
  pages={1782--1791},
  year={2025}
}

@inproceedings{nobis2019crfnet,
  title = {A Deep Learning-based Radar and Camera Sensor Fusion Architecture for Object Detection},
  author = {Nobis, Felix and Geisslinger, Maximilian and Weber, Markus and Betz, Johannes and Lienkamp, Markus},
  booktitle = {2019 Symposium on Sensor Data Fusion: Trends, Solutions, Applications (SDF)},
  year = {2019},
  pages = {1--7},
  doi = {10.1109/SDF.2019.8916629}
}

@inproceedings{nabati2021centerfusion,
  title = {CenterFusion: Center-based Radar and Camera Fusion for 3D Object Detection},
  author = {Nabati, Ramin and Qi, Hairong},
  booktitle = {Proceedings of the IEEE/CVF Winter Conference on Applications of Computer Vision (WACV)},
  year = {2021},
  pages = {1527--1536}
}

@ARTICLE{mssf2024,
  author={Liu, Hongsi and Liu, Jun and Jiang, Guangfeng and Jin, Xin},
  journal={IEEE Transactions on Intelligent Transportation Systems}, 
  title={MSSF: A 4D Radar and Camera Fusion Framework With Multi-Stage Sampling for 3D Object Detection in Autonomous Driving}, 
  year={2025},
  volume={26},
  number={6},
  pages={8641-8656},
  doi={10.1109/TITS.2025.3554313}}

@ARTICLE{xiong2024lxl,
  author={Xiong, Weiyi and Liu, Jianan and Huang, Tao and Han, Qing-Long and Xia, Yuxuan and Zhu, Bing},
  title={LXL: LiDAR Excluded Lean 3D Object Detection With 4D Imaging Radar and Camera Fusion}, 
  journal={IEEE Transactions on Intelligent Vehicles},
  volume={9},
  number={1},
  pages={79-92},
  year={2024},
  doi={10.1109/TIV.2023.3321240}
}

@INPROCEEDINGS{wang2022interfusion,
  author={Wang, Li and Zhang, Xinyu and Xv, Baowei and Zhang, Jinzhao and Fu, Rong and Wang, Xiaoyu and Zhu, Lei and Ren, Haibing and Lu, Pingping and Li, Jun and Liu, Huaping},
  booktitle={2022 IEEE/RSJ International Conference on Intelligent Robots and Systems (IROS)}, 
  title={InterFusion: Interaction-based 4D Radar and LiDAR Fusion for 3D Object Detection}, 
  year={2022},
  volume={},
  number={},
  pages={12247-12253},
  doi={10.1109/IROS47612.2022.9982123}}

@ARTICLE{xu2022rcfusion,
  author={Zheng, Lianqing and Li, Sen and Tan, Bin and Yang, Long and Chen, Sihan and Huang, Libo and Bai, Jie and Zhu, Xichan and Ma, Zhixiong},
  journal={IEEE Transactions on Instrumentation and Measurement}, 
  title={RCFusion: Fusing 4-D Radar and Camera With Bird’s-Eye View Features for 3-D Object Detection}, 
  year={2023},
  volume={72},
  number={},
  pages={1-14},
  doi={10.1109/TIM.2023.3280525}}

@inproceedings{kim2022craft,
  title={Craft: Camera-radar 3d object detection with spatio-contextual fusion transformer},
  author={Kim, Youngseok and Kim, Sanmin and Choi, Jun Won and Kum, Dongsuk},
  booktitle={Proceedings of the AAAI Conference on Artificial Intelligence},
  volume={37},
  number={1},
  pages={1160--1168},
  year={2023}
}

@ARTICLE{ramp_cnn,
  author={Gao, Xiangyu and Xing, Guanbin and Roy, Sumit and Liu, Hui},
  journal={IEEE Sensors Journal}, 
  title={RAMP-CNN: A Novel Neural Network for Enhanced Automotive Radar Object Recognition}, 
  year={2021},
  volume={21},
  number={4},
  pages={5119-5132},
  doi={10.1109/JSEN.2020.3036047}}

@inproceedings{harley2022simple,
  title={Simple-bev: What really matters for multi-sensor bev perception?},
  author={Harley, Adam W and Fang, Zhaoyuan and Li, Jie and Ambrus, Rares and Fragkiadaki, Katerina},
  booktitle={2023 IEEE International Conference on Robotics and Automation (ICRA)},
  pages={2759--2765},
  year={2023},
  organization={IEEE}
}

@inproceedings{caesar2020nuscenes,
  title     = {nuScenes: A Multimodal Dataset for Autonomous Driving},
  author    = {Caesar, Holger and Bankiti, Varun and Lang, Alex H. and Vora, Sourabh and Liong, Venice Erin and Xu, Qiang and Krishnan, Anush and Pan, Yuning and Baldan, Giancarlo and Beijbom, Oscar},
  booktitle = {Proceedings of the IEEE/CVF Conference on Computer Vision and Pattern Recognition (CVPR)},
  year      = {2020},
  pages     = {11618--11628},
  doi       = {10.1109/CVPR42600.2020.01164}
}

@article{kim2022kradar,
  title={K-radar: 4d radar object detection for autonomous driving in various weather conditions},
  author={Paek, Dong-Hee and Kong, Seung-Hyun and Wijaya, Kevin Tirta},
  journal={Advances in Neural Information Processing Systems},
  volume={35},
  pages={3819--3829},
  year={2022}
}

@misc{fent2024mantruckscenesmultimodaldataset,
      title={MAN TruckScenes: A multimodal dataset for autonomous trucking in diverse conditions}, 
      author={Felix Fent and Fabian Kuttenreich and Florian Ruch and Farija Rizwin and Stefan Juergens and Lorenz Lechermann and Christian Nissler and Andrea Perl and Ulrich Voll and Min Yan and Markus Lienkamp},
      year={2024},
      eprint={2407.07462},
      archivePrefix={arXiv},
      primaryClass={cs.CV},
      url={https://arxiv.org/abs/2407.07462}, 
}

@article{yang2025v2xradarmultimodaldataset4d,
  title={V2X-Radar: A Multi-modal Dataset with 4D Radar for Cooperative Perception},
  author={Yang, Lei and Zhang, Xinyu and Li, Jun and Wang, Chen and Ma, Jiaqi and Song, Zhiying and Zhao, Tong and Song, Ziying and Wang, Li and Zhou, Mo and Shen, Yang and Lv, Chen},
  journal={Advances in Neural Information Processing Systems (NeurIPS)},
  year={2025}
}

@inproceedings{yang2020radarnetexploitingradarrobust,
  title={Radarnet: Exploiting radar for robust perception of dynamic objects},
  author={Yang, Bin and Guo, Runsheng and Liang, Ming and Casas, Sergio and Urtasun, Raquel},
  booktitle={European conference on computer vision},
  pages={496--512},
  year={2020},
  organization={Springer}
}

@inproceedings{doppdrive2025,
  title={DoppDrive: Doppler-Driven Temporal Aggregation for Improved Radar Object Detection},
  author={Haitman, Yuval and Bialer, Oded},
  booktitle={Proceedings of the IEEE/CVF International Conference on Computer Vision},
  pages={26085--26094},
  year={2025}
}

@ARTICLE{VoD,
  author={Palffy, Andras and Pool, Ewoud and Baratam, Srimannarayana and Kooij, Julian F. P. and Gavrila, Dariu M.},
  journal={IEEE Robotics and Automation Letters}, 
  title={Multi-Class Road User Detection With 3+1D Radar in the View-of-Delft Dataset}, 
  year={2022},
  volume={7},
  number={2},
  pages={4961-4968},
  doi={10.1109/LRA.2022.3147324}}
}


\end{document}